\documentclass[runningheads]{llncs}
\usepackage{graphicx}
\begin{document}
\title{SocialBERT - Transformers for Online Social Network Language Modelling}
%
%
\author{Ilia Karpov\inst{1}\orcidID{0000-0002-8106-9426} \and
Nick Kartashev\inst{1}\orcidID{0000-0002-9722-3889} }

\authorrunning{Ilia Karpov \and Nick Kartashev}
\titlerunning{SocialBERT - Transformers for Online Social Network Language Modelling}
\institute{National Research University Higher School of Economics, Moscow, Russian Federation
\email{\{karpovilia, nickkartashev\}@gmail.com}}
%
%
%
\maketitle              

\begin{abstract}
The ubiquity of the contemporary language understanding tasks gives relevance to the development of generalized, yet highly efficient models that utilize all knowledge, provided by the data source. In this work, we present SocialBERT - the first model that uses knowledge about the author's position in the network during text analysis. We investigate possible models for learning social network information and successfully inject it into the baseline BERT model. The evaluation shows that embedding this information maintains a good generalization, with an increase in the quality of the probabilistic model for the given author up to 7.5\%. The proposed model has been trained on the majority of groups for the chosen social network, and still able to work with previously unknown groups. The obtained model as well as the code of our experiments is available for download and use in applied tasks
\end{abstract}

\section*{Introduction}
Online Social Networks (OSN) texts corpora size is comparable with the largest journalism, fiction, and scientific corpora. Evaluations within computational linguistics conferences and Kaggle competitions prove the feasibility of their automatic analysis. Traditional text processing tasks like morphological analysis, sentiment analysis, spelling correction are highly challenging in such texts. As a rule, by analyzing OSN texts we can observe a decrease in most quality metrics by 2-7\%. For instance, the best result on the sentiment detection on Twitter dataset at SemEval 2017 \cite{semeval2017} has an accuracy of 65.15, while a year earlier on the track SemEval 2016 \cite{semeval2016} the best result for the general English language has an accuracy of 88.13. Regarding the Russian language, comparison within the competition of morphological analysis tools MorphoRuEval-2017 \cite{MorphoRuEval2017} shows that the same tools work worse on the texts of online social networks than on fiction and news corpus - the lemmatization accuracy best result for OSN dataset is 92.29, while best result for literature dataset is 94.16.

Text processing quality decreasement is usually caused by a great amount of slang, spelling errors, region- and theme-specific features of such texts. This can be explained by the specifics of these texts being written by non-professionals, i.e., by the authors without the journalist education who have no opportunity or need for professional editing of their texts. Existing research also indicates the specifics of the social network communication itself, such as the tendency to transform oral speech to written text (orality), tendency to express emotions in written texts (compensation), and tendency to reduce typing time (language economy) \cite{crystal:2006}.

BERT \cite{devlin:19}, and its improvements to natural language modeling, which apply to extremely large datasets and sophisticated training schemes, solves the problems above to a great extent, by taking into consideration the corpora vocabulary at the pre-training step, and providing knowledge transfer from other resources at the fine-tuning step. For instance, Nguyen reports successful application of the RoBERTa model to OSN texts of Twitter users \cite{BERTweet}. Nevertheless, there is no consistent approach for analyzing social network users’ texts and no effective generalized language models that utilize the structure of OSN.

Unlike many other text sources, any social network text has an explicitly identifiable and publicly accessible author. This leads to a model that processes such a text, taking into consideration the characteristic features of its author. Such a model would make text analysis depending on the author's profile, significantly simplifying such tasks like correction of typos or word disambiguation by taking into account thematic interests and author's speech characteristics.

Thus, our objective is to define author latent language characteristics that capture language homophily. The homophily principle stipulates that authors with similar interests are more likely to be connected by social ties. The principle was introduced in the paper "Homogeneity in confiding relations" \cite{homogenity1}, by Peter Marsden. An interest’s homophily analysis can be found in the Lada A. Adamic paper on U.S. Elections \cite{adamic}. It shows that people with similar political views tend to make friends with each other. Usually, online social network users simultaneously have several interests and tend toward network-based homophily only with respect to some projection, such as political views, as shown by Lada Adamic. At the same time, online social network groups do not require any projections and are preferable for language structure modeling due to the following characteristics:
\begin{itemize}
    \item Groups and public pages (hereinafter “groups”) have their own pages. Texts, posted at group pages are mostly monothematic since group users are sharing the same interest or discussing news, important for a certain geographical region. In both cases, it is possible to identify group's specific vocabulary and speech patterns.
    \item The number of groups is two orders less than that of the users. This enables us to train a language model suitable for the entire online social network, without significant node filtering and computation costs.
    \item Groups generate a major part of text content, whereas many social network users do not write a single word for years because they act only as content consumers. At the same time, users’ interests are rather easily expressed through the groups they are subscribed to. 
\end{itemize}

Due to the reasons above, in this paper we focus on generating a group language model and keep the user language model outside the scope of this paper. Hereinafter we interpret a group as an author of texts written in this group's account. In the absence of explicit group attributes like age and gender, we focus on group homophilous relationships. We model groups social homophily through common users intersection to encourage groups with shared social neighborhoods to have similar language models.

In this paper we focused on the masked language modelling (MLM) task because, as a result of training such a model, one can create a better basic model for the analysis of social network texts which may be further adapted to the applied tasks listed above. We will discuss the effect of the basic model on the applied tasks in the Results section. 

Our key contributions are as follows:

\begin{itemize}
    \item We have generated a network embedding model, describing each group containing $5,000$ and more members. Observed groups have no topic limitations, so we can say that our training has covered all currently existing themes in the OSN, assuming that if some topic does not have at least one group with $5,000$ subscribers, then it is not important enough for language modeling. This can potentially lead to the model's inability to take into account highly specialized communities and tiny regional agglomerations, but does not affect the main hypothesis that author dependent language modeling can be more effective.
    \item We have proposed several new BERT-based models that can be simultaneously trained with respect to the group embedding, and performed training of an MLM-task using the group texts. Our best model achieved ~7.5\% perplexity increasement in comparison to the basic BERT model training for the same text corpora. This proves the appropriateness of the chosen approach. 
\end{itemize}

The rest of this paper proceeds as follows. Section 2 summarizes the related work on the modelling of OSN authors as network nodes and the existing approaches to language modelling and knowledge injection. Section 3 presents our proposed approach to continuous MLM with respect to network embedding. Section 4 presents details of the experimental setup, including the description of data collection and model training hyperparameters. We present the experimental results in Section 5 before making conclusions. 

\section*{Related Work}
In this section, we discuss related work on network node description and language modelling.

\subsection*{Author as Network Node}

The idea to use the author’s demographic features in order to improve the analysis quality had been offered before transformer based models were applied. The existing research e.g. of political preferences on Twitter \cite{adamic} or comments on Facebook \cite{netvizz}, proves the users’ inclination to establish relations with users with similar interests. 
In this work we want to model online social network group structure and language. Given users and groups simultaneously interact in OSN, we can use bipartite graphs to describe groups by their users and vice versa. For $user \rightarrow community$ bipartite graph, the affinity of groups may be described by number of common users: the more common subscribers they have, the greater their similarity is. Therein, various metrics can be applied such as correlation, for instance the Jaccard coefficient, cosine similarity etc. After calculating pairwise distances, one can obtain an adjacency matrix between all groups of the network. In order to reduce its dimensionality, methods based on random walk \cite{KEIKHA201847} and autoencoder models such as Deep Walk\cite{perozzi:2014}, Node2Vec\cite{node2vec} or a matrix factorization algorithm like NetMF\cite{netmf} may be used. Attention models may also be used for social representations. They are GraphBERT \cite{graphbert} or Graph Attention Networks\cite{gat}, but they are much more computationally expensive, and for this reason their use is limited for graphs of over ~$10^5$ node degree.

\subsection*{Language Modelling}

To the best of our knowledge, at the time of writing this paper, there is no published approach to the injection of online social network structure inside transformer-based deep learning language models. Various themes in the network may be considered standalone domains. Thus, the researches applying domain-specific adaptation become relevant \cite{gururangan2020dont}, \cite{han2019unsupervised}. These studies show that domain-relevant data is useful for training with both excessive and low resource problems. Since we want to develop single continuous model for all possible topics inside OSN, those approaches are significantly leveraged by a large (about 100-300) number of topics, depending on the granularity degree and absence of distinct borders between the social network communities.

Multi-domain adaptation \cite{yogatama2019learning} mechanisms applied in computer-based translation \cite{mghabbar2020building} are also of interest. Use of knowledge distillation \cite{hinton2015distilling} in training produces a positive result when there are split domains and their number is rather small. It makes the development of the continuous domain adaptation model relevant. 

\subsection*{Knowledge Injection}
One of the existing way of enhancing existing deep learning architectures is based on the knowledge injection approach. An example of graph data injection in BERT is the work of VGCN-BERT \cite{VGCN}, which adds graph information as a null token. This approach is similar to the first of our two proposed methods. The difference is that Lu proposes the addition of an ontological graph rather than a social network graph. Another approach, based on inserting additional layers to BERT model, is provided by Lauscher \cite{gurevych}. Authors show accuracy increasement up to 3\% on some datasets, simultaneously having the same or lower accuracy on other datasets. Our second model also modifies one of the BERT layers, but the proposed injection architecture is quite different.

\begin{figure*}[!t]
\begin{equation} 
\label{eq_cos}
(X_{v_i}, X_{v_k}) = \sum\limits_{j \in A \cap B} \frac{(1 - \frac{c_j}{M})^2}{c_j - \frac{c_j^2}{M}} + \sum\limits_{j \in \Omega \backslash (A \cup B)} \frac{(\frac{c_j}{M})^2}{c_j - \frac{c_j^2}{M}} + \sum\limits_{j \in (A \bigoplus B)} \frac{(\frac{c_j}{M} - \frac{c_j^2}{M^2})}{c_j - \frac{c_j^2}{M}}  
\end{equation}
\end{figure*}

\section*{Proposed Approach}

The proposed model takes into consideration the characteristics of a domain using a pre-computed social vector for the analysis of each token of incoming text. The general training process is as follows:
\begin{itemize}
    \item Generating adjacency matrices on the basis of network data – matrices preparation to evaluate the adjacency of two groups, based on mutual group members.
    \item Learning social vectors - obtaining the author’s vectors using factorization and random walk algorithms.
    \item BERT training, given pre-trained social vectors
\end{itemize}

\subsection*{Adjacency Matrix Generation}

When computing the social vector, we intended to have the opportunity to use the information on the community’s local environment as well as a description of its global position relative to all groups. To simulate local context we have chosen the DeepWalk algorithm. To capture the structure of our social graphs on a more global level, we used factorization of different kinds of pairwise distance matrices between the groups.

To calculate pairwise intersection sizes for our set of groups, we created a multithreaded C++ library, which yields an intersection matrix which, as shown later, is transformed into one of the various adjacency metrics.

\subsection*{Correlation coefficient}
\label{correlation}

Our first algorithm was based on factorizing a pairwise correlation matrix of our set of groups. Given a group as a vector of zeros and ones, having a length equal to the total number of users in the social network $N$, and containing a 1 for users who subscribed to our group, and a 0 otherwise. So, in this model, we represent the set of groups of size $M$ as a set of vectors $X_{v_i}$, each containing a sampling of a Bernoulli distribution. Without physical uploading of all vectors in our RAM, due to huge size of the resulting matrix, we calculated the sample correlation of our vectors based only on these easily computable variables: For set $A$ of subscribers of the group $a$, and the set $B$ of subscribers of the group $b$, we, as described in a previous section, calculated the intersection size $|A \cap B|$. Then, using this equation we obtain the correlation coefficient that will be used as one of the possible group affinity variables:

\begin{equation} 
\label{eq_cor}
cor(a, b) = \frac{|A \cap B|\cdot N - |A| \cdot |B|}{\sqrt{|A| \cdot |B| \cdot (N - |A|) \cdot (N - |B|)}}
\end{equation}

\subsection*{Cosine coefficient}

The most important difference of using cosine similarity instead of correlation as a distance metric between our groups, is that we normalized each user’s subscription string, therefore lowering the effect that users with a higher subscription count have on the resulting matrix.

Supposing that user $j$ is the member of $c_j$ groups. First of all, we need to subtract the mean from the respective row in our matrix of vectors $X_{v_i}$. After this transformation we will have $1 - \frac{c_j}{M}$ for positive subscription position and $-\frac{c_j}{M}$ for negative. We then need to divide each value by the standard deviation of a row. Dispersion equals $\frac{c_j^2}{M} + c_j - 2\frac{c_j^2}{M} = c_j - \frac{c_j^2}{M}$.

Therefore, after this transformation we will have value $(1 - \frac{c_j}{M}) / \sqrt{c_j - \frac{c_j^2}{M}}$ for positive subscription indicator and $-\frac{c_j}{M} / \sqrt{c_j - \frac{c_j^2}{M}}$ for negative.

So, the final expression will be as shown in equation \ref{eq_cos}, where $c_j$ denotes the count of subscriptions from user $j$. $M$ denotes the total number of groups. $A$ denotes set of subscribers of group $v_k$, $B$ denotes set of subscribers of group $v_i$. $X_{v_i}$ is a normalized vector for group $v_i$, and $X_{v_k}$ is a normalized vector for group $v_k$. $\Omega$ denotes set of users of the social network, and $\bigoplus$ denotes the symmetric difference between two sets. This formula depicts one of the metrics we used to calculate the similarity between two groups on a scale from -1 to 1.

\begin{figure*}
\centering
\includegraphics[width=.95\linewidth]{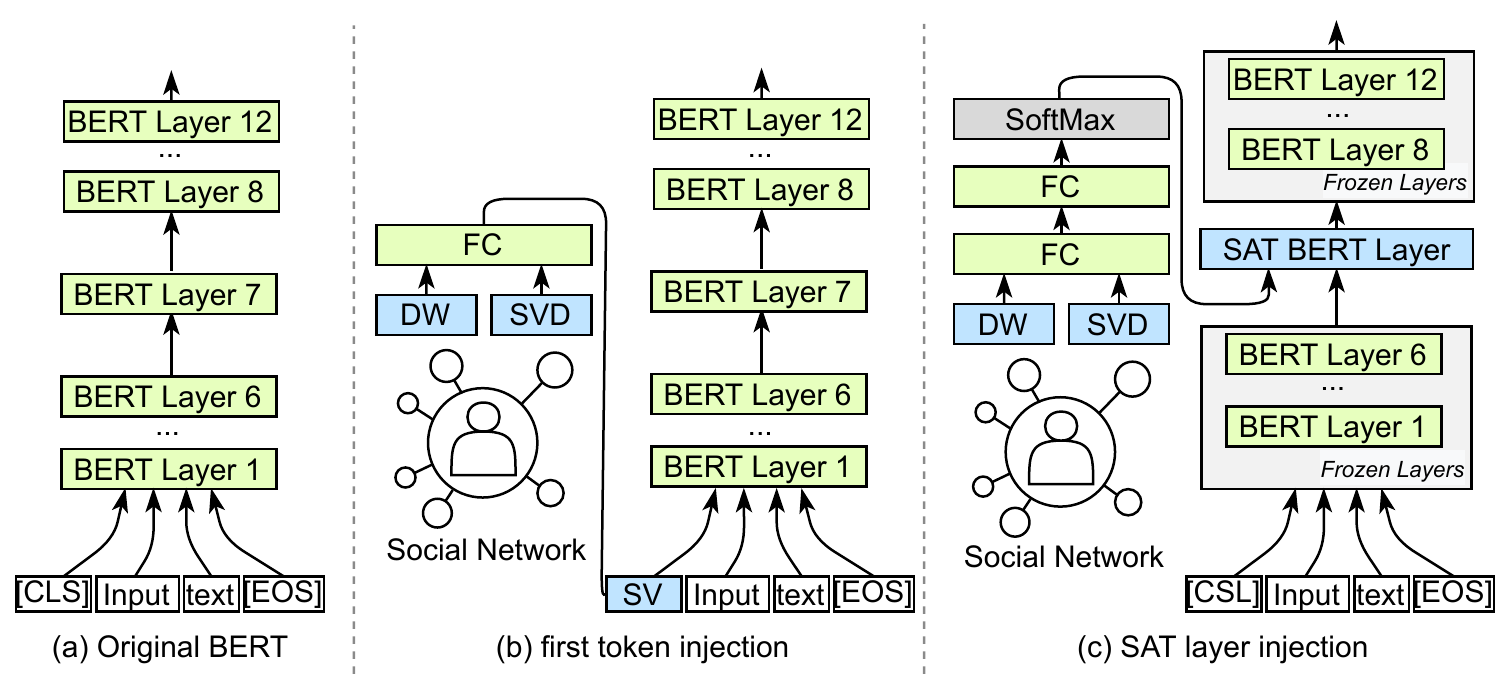}
\caption{Social vector injection methods}
\label{fig_bert_1}
\end{figure*}

\subsection*{Matrix Factorization}

We use a truncated SVD algorithm to closely estimate the distance between our groups with a pairwise scalar product of our embedded vectors.

For pairwise distance matrix $A$, we compute $U, \Sigma, V = SVD(A)$, and then obtain our vectors as rows of matrix $U \cdot \sqrt{\Sigma}$

\subsection*{Random Walk}

Given the group membership data, we can describe the measure of their closeness based on the Jaccard coefficient, which normalizes the number of common members of two groups by their size.

\begin{equation} 
\label{eq_jac}
Jac(a, b) = \frac{|A \cap B|}{|A| + |B| - |A \cap B|}
\end{equation}

Such a metric can be efficiently used for a random walk since it describes the probability of a transition from group $A$ to group $B$ with the "common user" edge. The resulting walks were used to train the DeepWalk model with the parameters recommended by the authors: $\gamma= 80$, $t = 80$, $w = 10$

\subsection*{BERT Training}

The vectors obtained independently as a result of random walk and SVD were integrated into the existing BERT Base model. The main purpose of the training was to teach the model to pay attention to the network vector. Here we used several different ways of embedding:

\begin{itemize}
    \item adding of a special social vector which concatenates both characteristics at the beginning of each sequence (\textbf{Zero token} injection).
    \item adding special \textbf{Social ATtention} (SAT) layer at various positions of the existing BERT model as described below.
\end{itemize}

The general scheme of both approaches is shown at the Figure \ref{fig_bert_1}. 

\begin{figure*}
\centering
\includegraphics[width=.9\linewidth]{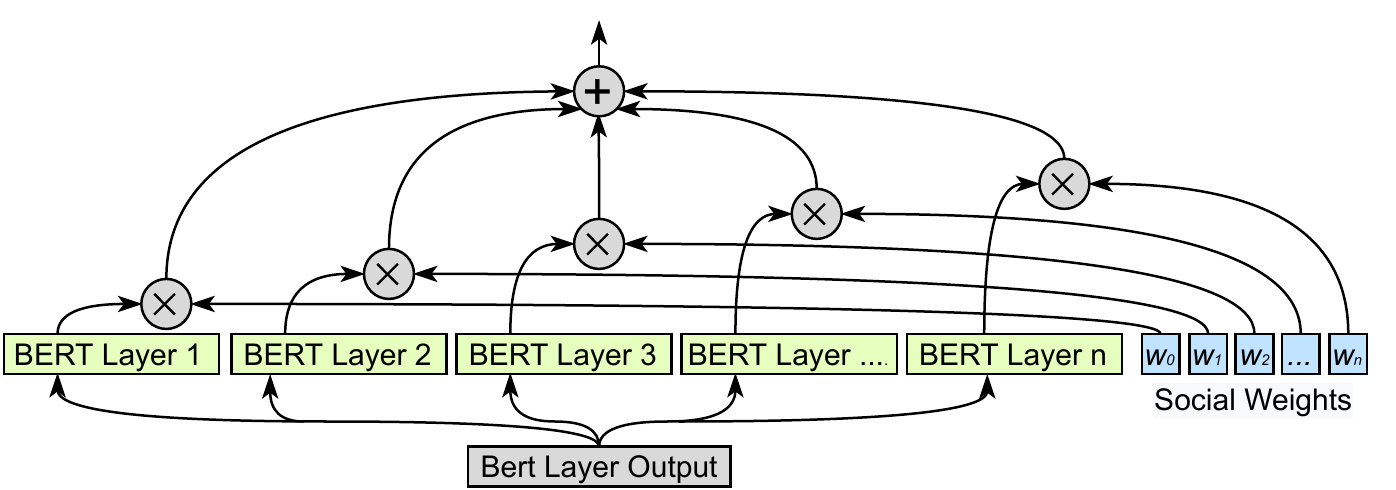}
\caption{Social Attention layer architecture}
\label{fig_bert_2}
\end{figure*}

To better inject social network information in our model, we created a special SAT layer. The injection mechanism depends on two hyperparameters: $i$ - number of BERT layer, chosen to be replaced by SAT layer, and $C$ - number of channels to use in our SAT layer. To inject Social Attention layer, first we pretrain basic BERT model on the entire training dataset for one epoch. Then, we freeze all layers of our model, and substitute i-th layer by our SAT layer, which shown in more detail at \ref{fig_bert_2}. The architecture of SAT layer is as follows: 

First, we build a 2-layer perceptron with GELU activation function between layers and SoftMax activation after second layer. We pass social network embeddings through that MLP thus obtaining new vectors $W$ with dimensionality reduced to $C$. Then, we create $C$ parallel BERT Layers, each initialised as substituted $i$-th layer of the original BERT. To compute output of SAT layer, we multiply each of the parallel bert layers with corresponding element of our resulting social vector $W$, and then summarise resulting vector sequences. The idea behind this method is to train each of our $C$ BERT layers to be responsive for a superset of social network topics, and then represent each author as a composition of this supersets.

\section*{Experimental Setup}
This section describes the parameters of the approach we have proposed, and varying hyperparameters of the trained models.

\subsection*{Data Collection}
We have used the social network VKontakte, which comprises 600 million users and 2.9 million groups. The majority of its users are Russian-speaking Internet users. The social network has a rich API that automatically provides a significant amount of data related to texts and network characteristics of the network nodes (community members, users’ friends). 

The VKontakte social network makes it possible to receive messages from groups, public pages, and event pages, through the program interface for non-commercial use\footnote{https://vk.com/dev/rules}. Storage and transmission of users' personal data, including the user's primary key (identifier), are restricted.

We performed sha3 hashing of all user identifiers during the data collection step. This operation makes it impossible to calculate precisely exact user’s membership in a community, while at the same time, preserving the bipartite graph structure. We did hashing with python-sha3 library\footnote{https://github.com/bjornedstrom/python-sha3}.

First we collected information on the size of common ties, then we selected the communities comprising $5,000$ or more members. We have established this threshold because small groups are often closed or updated irregularly. It lengthens the stage of the matrix preprocessing and does not improve the quality of the model training. Our final network included $309,710$ communities with given sizes.

We collected $1,000$ messages from 2019 for each selected group. The above period was chosen because we intended to obtain a thematic structure of the network which was not influenced by recent epidemiological issues. If a group wrote fewer than $1,000$ messages, we used the actual number of texts. If the community wrote more than the abovementioned number of messages, we randomly chose $1,000$ messages. The length of the majority of messages was less than 500 words. For the construction of the text language model, we used only the first 128 tokens of the text.

\subsection*{Network Modelling}
We performed DeepWalk \cite{perozzi:2014} training, applying the standard parameters recommended by the authors.

Independently, we managed to obtain our social vectors using DeepWalk and SVD of correlation matrices and cosine group-wise distance. To compute SVD part of social embedding computing, we were using fbpca\footnote{https://fbpca.readthedocs.io} framework by Facebook, with n\_iters = 300 and others parameters set by default, and which was running for 10 hours on 20 CPU cores (40 threads) Intel Xeon(R) Gold 5118 CPU with 1TB RAM.

\subsection*{Language Model Training}

As a base for our language modelling experiment, we used RuBERT: Multilingual BERT, adapted and pre-trained for the Russian language by DeepPavlov \cite{deeppavlov}. We ran a series of experiments to compare the ways in which social embeddings were integrated into BERT and the ways in which they were obtained.  

For the purpose of the training and the evaluation of our model, we created 3 different datasets, as to better illustrate the performance of our model in different situations. First dataset, containing posts from $278,739$ randomly chosen groups, is used by our model on the training stage, so it will be referenced as training set below. Second dataset, also containing posts from the same $278,739$ groups as in training, contains new text data, previously unseen by our model on the training stage. This dataset is called validation-known \textit{(val-k)}. The final datasets, contains posts from remaining $30,971$ groups, so when validating on this data the only information our model knows about the source is the social embeddings we pass to our model, which makes the task a little more challenging, because unlike in the validation-known dataset our model hasn't seen different posts from the same author on the training stage. This dataset is called validation-unknown \textit{(val-u)}.

From our initial set of $309,710$ groups we selected $189,496$ groups which contained at least 5 texts of sufficient length, with mean number of texts per group of $174.48$, and standard deviation of $125.5$.

All our experiments were conducted on a machine with Tesla V100 GPU with 32 GB of video memory for BERT training and Intel Xeon(R) Gold 5118 CPU with 1TB RAM for random walk and matrix factorization.

There was a total count of $43,232,000$ training sequences, $5,404,000$ val-k sequences, and $5,404,000$ \textit{val-u} sequences in our data.

Each experiment was trained for a total of one to two weeks on Tesla V100, in each experiment $1,351,000-2,702,000$ training steps were made (training was stopped in case of overfitting). 

Each experiment used a learning rate of 1e-5, and an Adam optimizer with a warmup of $20,000$ steps. Random seed was fixed for each series of experiments, and a total of 5 series of experiments with different random seeds were conducted. 

\begin{itemize}
    \item Social embedding vector added to zero token embedding, uses concatenation of vectors from SVD of correlation matrix and vectors obtained by DeepWalk.
    \item Social embedding vector added to zero token embedding, uses concatenation of vectors from SVD of cosine similarity matrix and vectors obtained by DeepWalk.
    \item Social embedding vector added to zero token embedding, uses just vectors obtained by DeepWalk.
    \item Baseline BERT, no social network embeddings used.
    \item BERT with SAT layer. It uses concatenation of vectors of SVD of the correlation matrix and the vectors obtained by DeepWalk as a social embedding. 
\end{itemize}

Throughout our experiments we found out that layer number hyperparameter $i$ has no significant meaning on our model performance, so we chose $i = 11$, as the best value for $i$ with insignificant lead. We found no improvement when increasing $C$ past 32, however, times and memory costs were very high, so we stopped with value $C = 32$.  

\section*{Results}

\begin{figure}
\centering
\includegraphics[width=.9\linewidth]{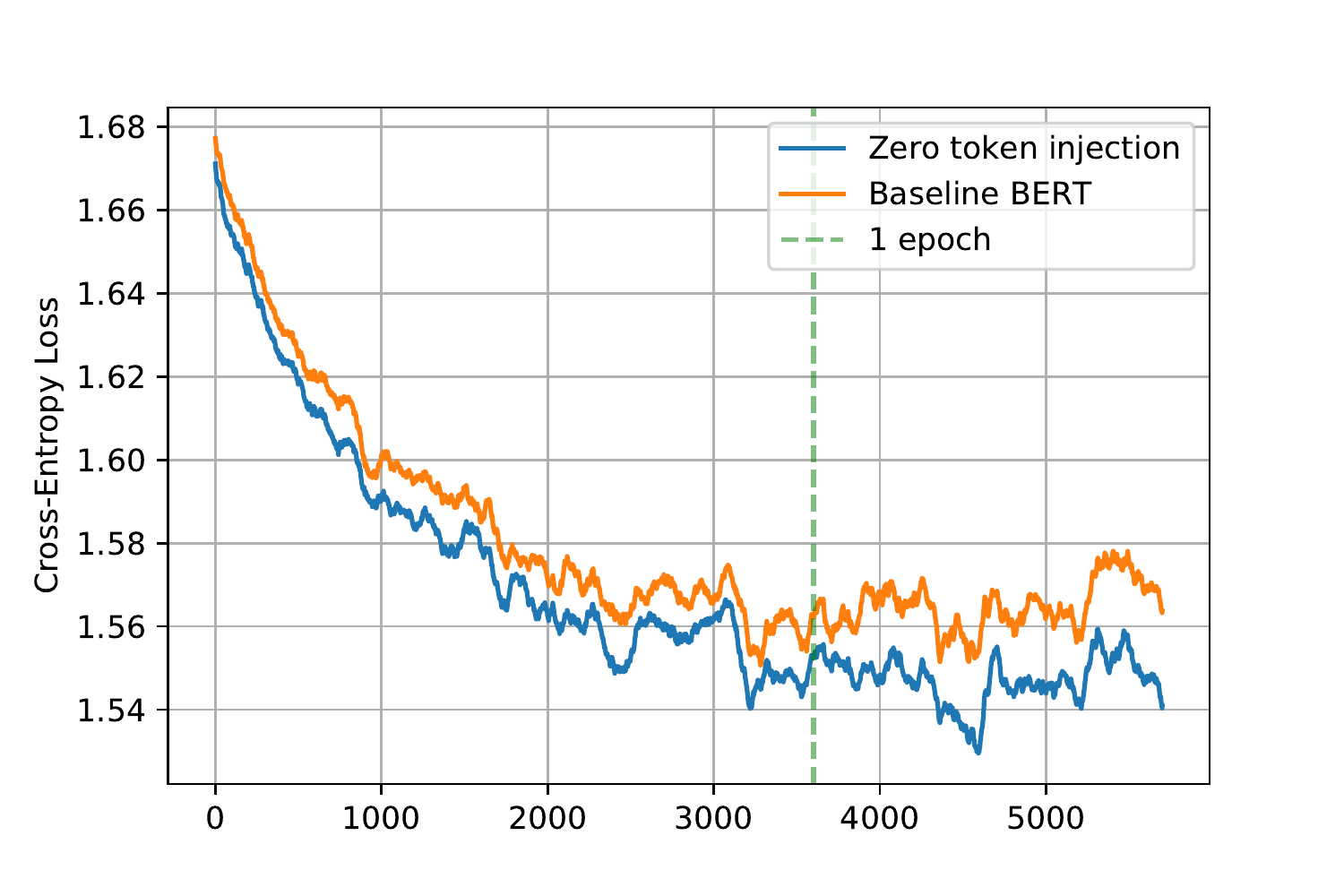}
\caption{Zero injection test evaluation, \textit{val-u} dataset}
\label{fig_test_1}
\end{figure}

\begin{figure}
\centering
\includegraphics[width=.9\linewidth]{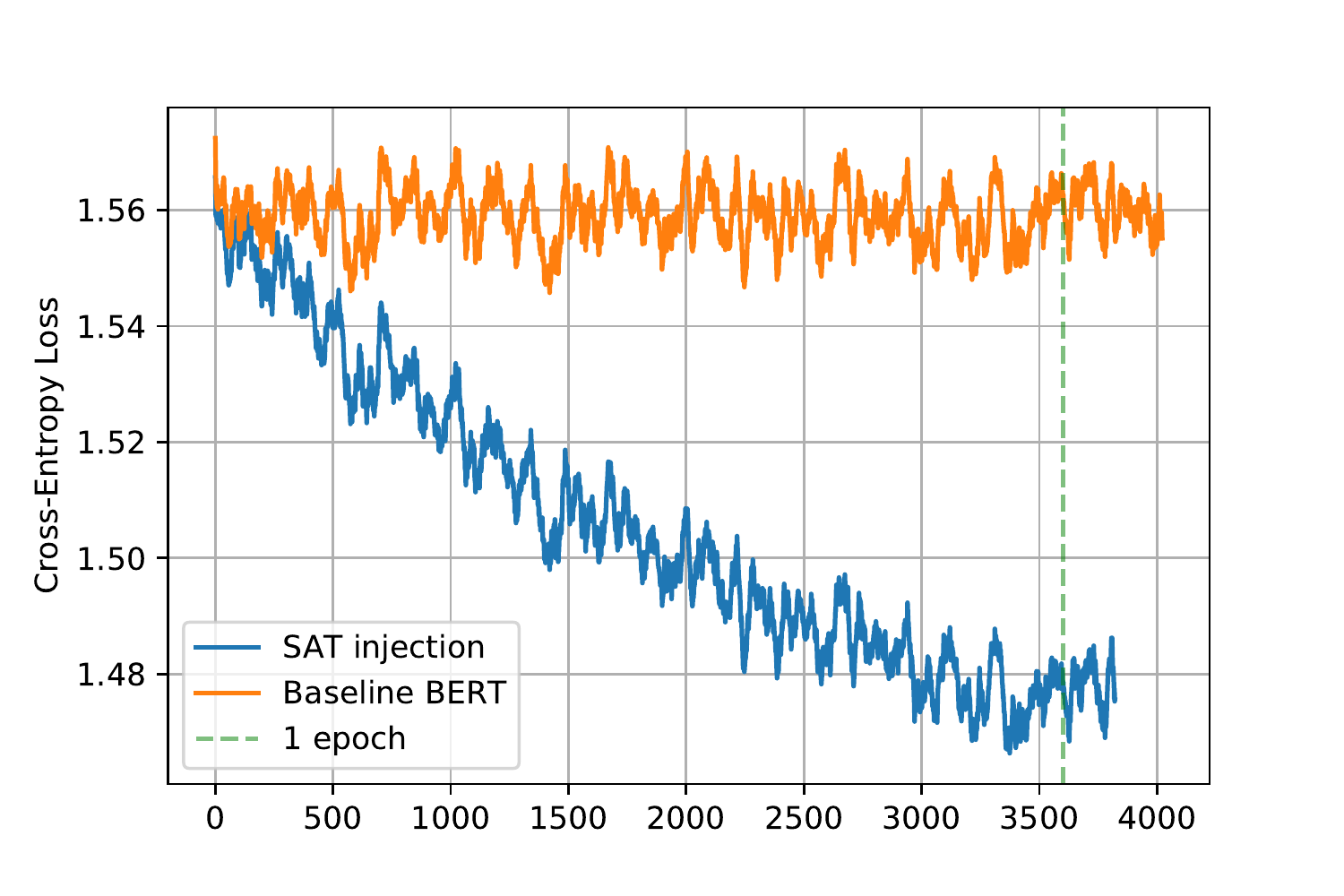}
\caption{SAT layer test evaluation on \textit{val-u} dataset}
\label{fig_test_2}
\end{figure}

We evaluated the obtained model using the quality of predicting the missing token in the sentence and the perplexity measure, used in the original works such as BERT and RoBERTa. The absolute value of perplexity for the given model depends on many parameters such as the size of the model vocabulary, tokenization parameters, and fine-tuning dataset. Thus, it is rather difficult to evaluate the direct perplexity influence on the solution of any applied problem. Our case is further complicated by the necessity to prepare our own benchmark, since, as far as we know, none of the existing datasets contain the information about the author’s social ties used by our model.

On the other hand, perplexity difference for the same basic model, trained on exactly the same corpus with the same preprocessing, must affect the quality of the applied tasks, as shown by the RoBERTa and original BERT paper authors: as perplexity decreases, the quality of classification on the SST-2 (for RoBERTa and BERT), MNLI-m and MRPC (for BERT)  dataset increases. Thus, the perplexity difference for two initially identical BERT models, trained on the same texts, indicates better trainability and further effectiveness for the model with lower perplexity.

The original BERT paper \cite{devlin:19} reports a perplexity of 3.23 for the 24 layer model with 1024 token input. The BERT Base model, trained on the same corpora has a perplexity of at least $3.99$ both for English and Russian language. This can be explained by the significant variation in topics, and even languages, covered by those models. Since online social network (OSN) texts are a subset of the entire text array, training only on OSN reduces perplexity to $2.83$ for the multilanguage BERT Base model (RuBert OSN). Further improvement is possible through the use of additional information regarding social vectors, allowing the evaluation measure to be reduced to $2.72$, as shown in Table \ref{AbsoluteQualityTable}.

\begin{table}
\centering
\begin{tabular}{|l|l|l|l|}
\hline \textbf{SNA Model} & \textbf{LM Model} & \textbf{Data} & \textbf{Loss} \\ \hline
 --- &  Baseline BERT & & $1.568$\\
 DW only &  Zero token inj. &  & $1.563$ \\
 Cos. \& DW & Zero token inj. & \textit{val-u} & $1.551$ \\
 Corr. \& DW &  Zero token inj. & & $1.542$  \\
 Corr. \& DW &  SAT injection &  &  \textbf{1.473} \\
 \hline
 ---   &  Baseline BERT &  & 1.500\\
 Corr. \& DW &  Zero token inj. & \textit{val-k} & $1.486$  \\
 Corr. \& DW &  SAT injection &  & \textbf{1.393} \\
\hline
\end{tabular}
\caption{\label{restable1} Comparison of various network vectors and strategies of BERT Pretraining }
\end{table}

Table \ref{restable1} shows the averaged loss function for last 50 iterations before the model stops training. The best result is achieved when using the concatenation of the Deep Walk (DW) embedding and the correlation coefficient (Corr.) as the network vector. Concatenation of cosine similarity and Deep Walk (DW) shows a bit worse results. Validation of model on \textit{val-u} dataset (Figure \ref{fig_test_1}) shows that injecting (Corr. \& DW) network vector into a zero token improves the base BERT model by no more than $0.03$ points of loss function. Further training doesn't lead to any improvements. Replacing the eleventh layer of the BERT model with the SAT layer improves the model by $0.21$ points compared to the baseline BERT loss results (Figure \ref{fig_test_2}). 

Figures \ref{fig_test_1} and \ref{fig_test_2} are built for the unknown texts of earlier unknown groups (dataset \textit{val-u}). Evaluation on the unknown texts of the known groups (dataset \textit{val-k}) shows more significant increase up to $0.11$ points of loss function. Most of the groups we have selected do not change subscribers and topic significantly over time, which will allow either, to use pretrained group embedding groups for analysis, or to search for the most similar community, based on social network characteristics. 

\begin{table}
\centering
\begin{tabular}{|l|l|l|l|}
\hline \textbf{Model} & \textbf{Perplexity} & \textbf{Loss} \\ \hline
BERT 12L & 3.54 & 1.82 \\
BERT Large & 3.23 & 1.69  \\
RuBert & 4.0 & 2.00 \\
RuBert OSN & 2.83 & 1.50 \\
SocialBERT & \textbf{2.62} & \textbf{1.39}\\
\hline
\end{tabular}
\caption{\label{AbsoluteQualityTable} BERT base models perplexity}
\end{table}

Table \ref{AbsoluteQualityTable} shows relative difference of perplexity and loss function for different forks of initial BERT Base model. 
We can observe that a two times increase in number of BERT layers can reduce perplexity by 8.8\% from 3.54 to 3.23. At the same time, the use of network vectors can reduce perplexity by 7.5\% from $2.83$ to $2.62$ for the \textit{val-k} dataset. This result is comparable to the 8\% perplexity improvement within the RoBERTa model.

We consider the proposed model useful for all language understanding tasks that implicitly use probabilistic language modeling, first of all, entity linking, spell-checking, and fact extraction. The model shows very promising results on short messages and texts with poor context:

(1) The obtained examples demonstrate that the model successfully learns regional specifics. For example, for the \textit{"[MASK] embankment"} pattern, the basic BERT model recommendation is \textit{"Autumn embankment"}, while the model initialized with the Saint Petersburg regional groups offers \textit{"\textbf{Nevskaya} embankment"} based on the Neva River in the Saint Petersburg.

(2) The model can be useful for Link Prediction tasks on short texts. For example, for the pattern \textit{"we read Alexander [MASK] today"} baseline BERT model returns \textit{"we read Alexander Korolev today"} (actor and producer) while model with poetry group vector initialization returns \textit{"we read Alexander \textbf{Blok} today"} (well known poet) .

(3) It is also useful in tasks of professional slang detection. For example, given the pattern "Big [MASK]", basic BERT model returns \textit{"Big \textbf{bro}"} while model with Data Science group vector returns \textit{"Big \textbf{data}"}.

\section*{Conclusion}
In this paper we present the SocialBERT model for the author aware language modelling. Our model injects the author social network profile to BERT, thus turning BERT to be author- or domain- aware. 

We select ~ 310 thousand groups and ~ 43 million texts that describe nearly all topics being discussed in the entire network. The proposed model demonstrates its effectiveness by improving the value of perplexity for the Masked Language Modelling task by up to 7.5\%. The model has the best results for new texts of already seen groups, still showing good transfer learning for texts of earlier unseen groups. We believe that the proposed model can be useful as a basic model for text analysis of Online Social Network texts and lead to author-aware generative models.

\section*{Acknowledgements}
The article was prepared within the framework of the HSE University Basic Research Program and through computational resources of HPC facilities provided by NRU HSE.

\bibliographystyle{unsrt}
\bibliography{sample}

\end{document}